\documentclass[preprint]{elsarticle}

\usepackage{amsmath,amssymb,graphicx}
\usepackage[colorlinks=true]{hyperref}
\usepackage{verbatim}
\usepackage[utf8]{inputenc}
\journal{Computer Methods and Programs in Biomedicine}

\begin{document}
\begin{frontmatter}

\title{LM-CartSeg: Automated Segmentation of Lateral and Medial Cartilage and Subchondral Bone for Radiomics Analysis}

\author[1]{Tongxu Zhang}
\author[1,2]{Zongpan Li\corref{cor2}}
\author[1]{Aaron Kam Lun Leung\corref{cor1}}
\author[1]{Siu Ngor Fu}
\cortext[cor1]{Corresponding author}
\cortext[cor2]{The jobs was completed during his student years at The Hong Kong Polytechnic University}

\address[1]{Department of Rehabilitation Sciences, The Hong Kong Polytechnic University, Kowloon, Hong Kong, China}
\address[2]{Department of Physical Therapy and Rehabilitation Science, University of Maryland, Baltimore, Maryland, USA}

\begin{abstract}
\textbf{Background and Objective:}
Radiomics of knee MRI requires robust, anatomically meaningful regions
of interest (ROIs) that jointly capture cartilage and subchondral bone.
Most existing work relies on manual ROIs and rarely reports quality
control (QC). We present LM-CartSeg, a fully automatic pipeline for
cartilage/bone segmentation, geometric lateral/medial (L/M)
compartmentalisation and radiomics analysis.

\textbf{Methods:}
Two 3D nnU-Net models were trained on SKM-TEA (138 knees) and
OAIZIB-CM (404 knees). At test time, zero-shot predictions were fused
and refined by simple geometric rules: connected-component cleaning,
construction of 10\,mm subchondral bone bands in physical space, and a
data-driven tibial L/M split based on PCA and $k$-means. Segmentation
was evaluated on an OAIZIB-CM test set (103 knees) and on SKI-10
(100 knees). QC used volume and thickness signatures. From 10 ROIs we
extracted 4\,650 non-shape radiomic features to study inter-compartment
similarity, dependence on ROI size, and OA vs.\ non-OA classification on
OAIZIB-CM and a clinical Po-OA cohort (185 knees).

\textbf{Results:}
Post-processing improved macro ASSD on OAIZIB-CM from 2.63 to 0.36\,mm
and HD95 from 25.2 to 3.35\,mm, with DSC $\approx$0.91; zero-shot DSC on
SKI-10 was $\approx$ 0.80. The geometric L/M rule produced stable
compartments across datasets, whereas a direct L/M nnU-Net showed
domain-dependent side swaps. Only 6--12\% of features per ROI were
strongly correlated with volume or thickness. Radiomics-based models
achieved AUC up to 0.91 (OAIZIB-CM) and 0.83 (Po-OA), clearly exceeding
models restricted to size-linked features.

\textbf{Conclusions:}
LM-CartSeg yields automatic, QC'd ROIs and radiomic features that carry
discriminative information beyond simple morphometry, providing a
practical foundation for multi-centre knee OA radiomics studies. Code is available at \url{https://github.com/jukieCheung/LM-CartSeg}.
\end{abstract}



\end{frontmatter}
\section{Introduction}

Knee osteoarthritis (KOA) is one of the most prevalent musculoskeletal disorders worldwide and a leading cause of pain, disability, and loss of quality of life in older adults. Recent Global Burden of Disease analyses estimate that osteoarthritis affected roughly 595~million people in 2020, corresponding to about 7.6\% of the global population, with knee involvement accounting for a large and rapidly growing share of cases \cite{vina2018epidemiology, steinmetz2023global}. The incidence and prevalence of KOA continue to rise in parallel with population ageing and obesity, posing substantial socioeconomic and healthcare challenges.

\begin{figure*}[htbp!]
\centering
\includegraphics[width=\linewidth]{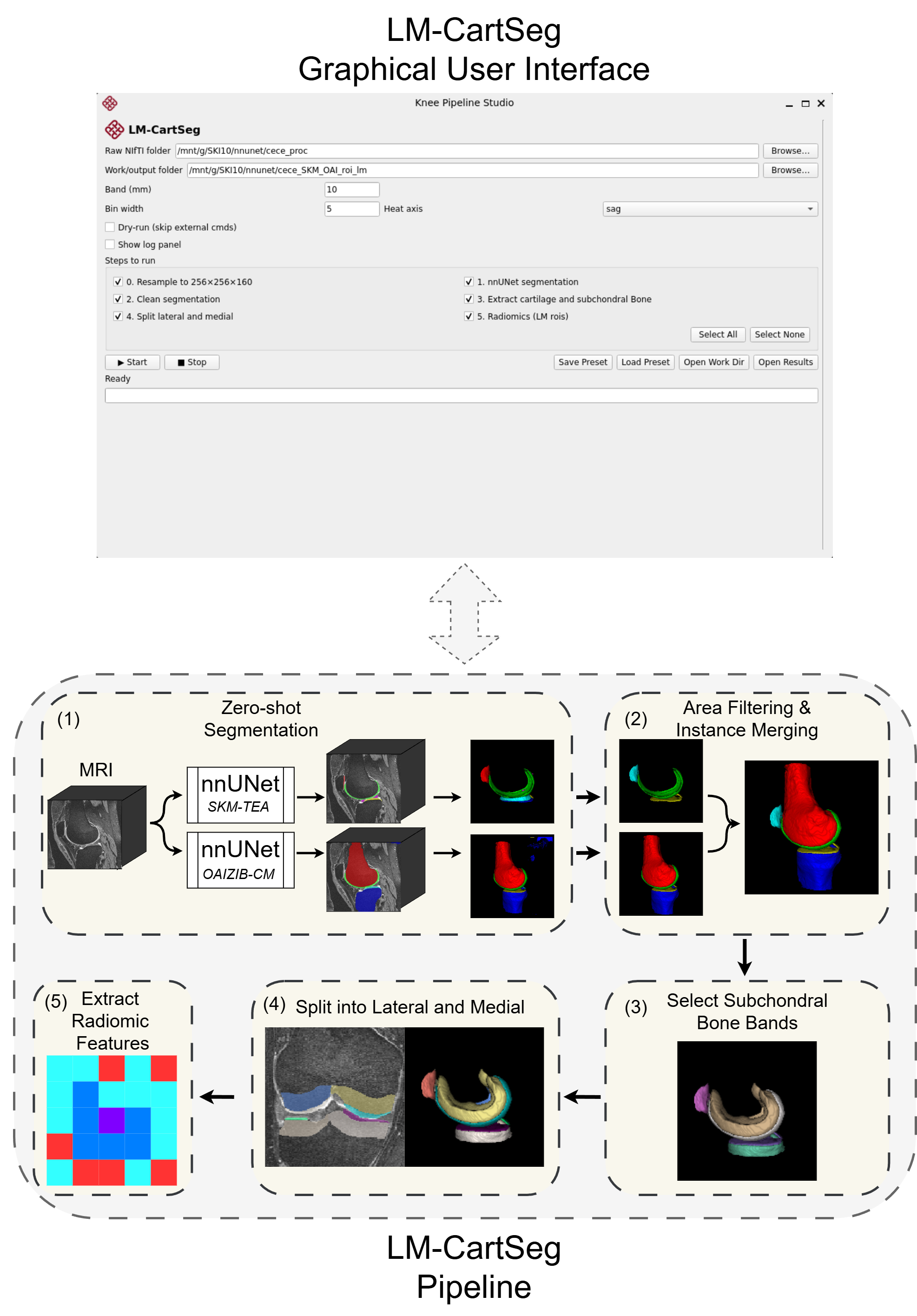}
\caption{Overview of the proposed pipeline: LM-CartSeg. A 3D knee MR volume is processed by two pretrained nnU-Net models (SKM-TEA and OAI/ZIB-CM) for zero-shot segmentation, followed by area filtering and instance merging to obtain clean cartilage and bone labels. Femoral and tibial cartilage are then split into lateral and medial compartments, subchondral bone bands are generated, and radiomic features are extracted for downstream analysis.}
\label{fig:1}
\end{figure*}

Magnetic resonance imaging (MRI) provides a three-dimensional, multi-tissue view of the knee joint, enabling direct assessment of articular cartilage, menisci, and subchondral bone \cite{de2021detection, wang2012use}. Quantitative sequences such as T\textsubscript{2} and T\textsubscript{1$\rho$} mapping can detect biochemical alterations in cartilage before radiographic damage appears, and have been associated with subsequent development or progression of KOA \cite{gomoll2011preoperative, cao2023application}.

Radiomics, the high-throughput extraction of large numbers of quantitative features from medical images---has been proposed as a bridge between medical imaging and precision medicine \cite{lambin2017radiomics, van2020radiomics}. By systematically quantifying intensity, texture, and shape patterns, radiomic features can capture subtle tissue alterations that are invisible to the naked eye and may carry prognostic or predictive information. In KOA, MRI-based radiomics frameworks have been developed to identify OA using features from cartilage, subchondral bone, and other joint tissues,\cite{hirvasniemi2021machine, xue2022radiomics, cui2023development, fu2025mri, chen2025integration, angelone2024knee} to predict structural or symptomatic progression, and to construct multi-tissue signatures that integrate cartilage and subchondral bone information.These studies demonstrate the potential of radiomics for early KOA assessment but also highlight important methodological challenges.

They main focus on the cartilage\cite{cui2023development, angelone2025innovative}, subchondral bone \cite{hirvasniemi2021machine, lin2023prediction} or both of them \cite{xie2021radiomics, xue2022radiomics}. However, most of these tasks require the involvement of doctors or radiologists in manual ROI segmentation. Therefore, it is necessary to propose an automatic segmentation of cartilage and subchondral bone to reduce human consumption. And this method can be used for lateral and medial cartilage, and the thickness of subchondral bone should be set according to specific requirements.

Recent advances in deep learning have enabled robust automatic segmentation in medical imaging \cite{isensee2021nnu, chadoulos2022novel, khan2022deep, isensee2024nnu, lv2025lmsst}, including applications to knee bones and cartilage on MRI. Ambellan et al.\ \cite{ambellan2019automated} introduced a framework that combines 3D statistical shape models with 2D/3D convolutional neural networks to segment femur, tibia, and cartilage compartments in Osteoarthritis Initiative (OAI) data, forming the basis of the OAIZIB-CM dataset. Subsequent work has extended these ideas to fully automatic cartilage subregional assessment and to heterogeneous clinical MRI protocols \cite{yao2024cartimorph}. However, most existing KOA radiomics studies \cite{wirth2010spatial, surowiec2014t2, panfilov2022deep} still analyze either cartilage or subchondral bone in isolation, often in a limited number of compartments, and rarely exploit fully automatic, anatomically detailed regions-of-interest (ROIs) that jointly cover lateral/medial (L/M) cartilage and the corresponding subchondral bone bands.

In this work, we build on automatic segmentation to construct an end-to-end, fully automated pipeline for knee radiomics analysis (Fig.~\ref{fig:1}). Starting from routine 3D knee MRI, we:
\begin{itemize}
  \item automatically segment femoral, tibial, and patellar cartilage in 3D;
  \item derive \emph{subchondral bone bands} of fixed physical thickness inside the femur and tibia, measured inward from the cartilage–bone interface;
  \item partition femoral/tibial cartilage and their subchondral bands into lateral (L) and medial (M) compartments using a purely geometric rule that generalizes across cohorts;
  \item extract standardized radiomics features from all cartilage and subchondral ROIs.
\end{itemize}

We systematically evaluate segmentation accuracy and the robustness of the L/M partition across three cohorts, and use simple morphometric quantities (volume, mean thickness, and compartmental ratios) as sanity checks and quality-control indicators. Within each cohort, we further compare OA classification performance between size-linked descriptors (volume/thickness) and texture radiomics to quantify the incremental value of texture under a realistic, fully automatic pipeline, rather than to build a cross-center deployable classifier.

\section{Methods}

\subsection{Network training and inference by nnU-Net}
\label{subsec:nnunet}

We adopt the 3D full-resolution nnU-Net framework the state-of-the-art segmentaion model in medical imaging\cite{isensee2021nnu, isensee2024nnu}
to segment femoral, tibial and patellar cartilage and the corresponding
bones. Two models are trained independently on the public SKM-TEA
\cite{desai2022skm} and OAIZIB-CM
\cite{ambellan2019automated,yao2024cartimorph} datasets, respectively,
using the provided manual labels as ground truth. The label sets differ
slightly: SKM-TEA provides femoral and patellar cartilage, bilateral
menisci and bilateral tibial cartilage, whereas OAIZIB-CM provides
femur, femoral cartilage, tibia and bilateral tibial cartilage.

For each dataset, nnU-Net uses its default preprocessing, resampling to
a dataset-specific target spacing and $z$-score intensity
normalization \cite{carre2020standardization, zwanenburg2020image, dovrou2023segmentation}. Also default data augmentations, like random flips, rotations,
scalings and intensity transforms. And the standard combined Dice and
cross-entropy loss. Among them, the hyperparameters for training, except for the epoch multiple set to 250 rounds, are automatically optimized by nnU-Net. Training and validation splits are defined at the subject level to avoid information leakage.

At test time, we run both trained models with the default nnU-Net
inference settings, such as sliding-window prediction with test-time
mirroring, and obtain one softmax probability map per model. These
probability maps are fused in probability space (voxelwise averaging),
and the final hard labels are obtained by voxelwise argmax followed by
connected-component filtering and anatomical instance merging.

\subsection{Geometric post-processing and explainable rules}
\label{subsec:postproc}

\begin{figure}
\centering
\includegraphics[width=\linewidth]{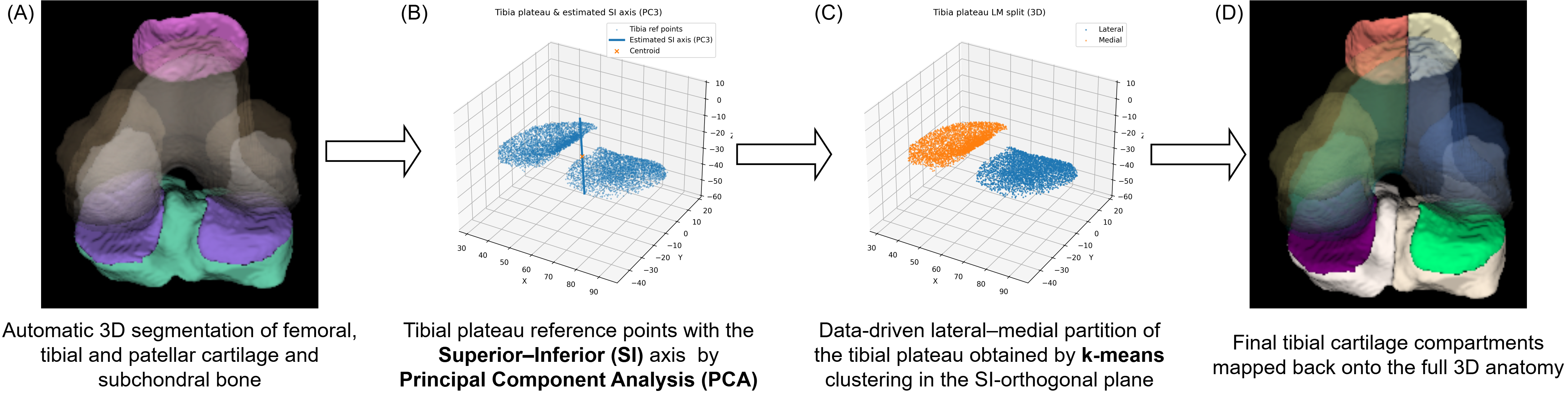}
\caption{Data-driven tibial compartment definition: (A) Segmented cartilage and subchondral bone,
(B) PCA \cite{hotelling1933analysis} derived superior–inferior axis,
(C) k-means \cite{macqueen1967multivariate} lateral–medial partition in the SI-orthogonal plane,
(D) Resulting tibial compartments on the 3D anatomy.}
\label{fig:2}
\end{figure}

Starting from the voxelwise nnU-Net predictions, we apply a sequence of
fully rule-based geometric post-processing steps to obtain anatomically
plausible, analysis-ready masks. This includes connected-component cleaning to remove obvious segmentation errors, construction of fixed-thickness subchondral
bone bands, and a data-driven yet anatomically interpretable partition of lateral (L) and medial (M) compartments, as shown in Fig. \ref{fig:2}.

\paragraph{Connected-component cleaning}
For each anatomical label, we compute 3D connected components (6-connectivity). We apply label-specific, interpretable rules:
\begin{itemize}
  \item \textbf{Bones:} keep the largest component near the volume center; optionally discard components touching the acquisition borders.
  \item \textbf{Cartilage:} keep all components with size $\ge$ a minimum threshold. Thresholds are specified in physical volume (mm$^3$) and converted to voxels via $n_{\min}=\lfloor V_{\min}/(s_x s_y s_z)\rfloor$, where $(s_x,s_y,s_z)$ are voxel sizes in mm.
\end{itemize}

\paragraph{Subchondral bone bands}
Let $B(x)\in\{0,1\}$ denote the bone mask (femur or tibia), $C(x)\in\{0,1\}$ the corresponding cartilage mask, and $\mathbf{s}=(s_x,s_y,s_z)$ the voxel spacing in mm. We compute the Euclidean distance transform (EDT) on the complement of cartilage with physical sampling:
\begin{equation}
  D(x) \;=\; \min_{y:\,C(y)=1} \,\big\| (x-y)\odot \mathbf{s} \big\|_2 .
\end{equation}
For a target band thickness $t\in\{5,10\}$\,mm, which could be set in the UI to choose others thickness, the subchondral band is defined as
\begin{equation}
  \Omega_{\text{sub}} \;=\; \big\{\, x \;\big|\; B(x)=1 \;\land\; D(x)\le t \,\big\}.
\end{equation}
This yields a geometrically meaningful layer restricted to bone and equidistant (in mm) from the cartilage interface.

\paragraph{Lateral/Medial (L/M) compartment partition}
For each knee we derive a case-specific L/M boundary directly from the tibial plateau geometry.
Let $\Omega_{\text{ref}}\subset\mathbb{Z}^3$ denote a tibial reference mask and
$\{x_i\}_{i=1}^N\subset\mathbb{R}^3$ the corresponding world-coordinate points obtained from the image affine.

\begin{enumerate}
  \item \textbf{Reference region.}
  We preferentially take $\Omega_{\text{ref}}$ as the tibial cartilage mask; if it contains fewer than $200$ voxels
  ($N<200$), we fall back to the tibial subchondral band. This yields a point cloud
  $X = \{x_i\}_{i=1}^N$ sampling the tibial plateau.

  \item \textbf{Local plane via PCA.}
  We compute the centroid
  $c = \frac{1}{N}\sum_{i=1}^N x_i$ and fit a 3D PCA to $X$.
  Let $(e_1,e_2,e_3)$ be the eigenvectors of the sample covariance matrix
  $\Sigma = \frac{1}{N}\sum_{i=1}^N (x_i-c)(x_i-c)^\top$,
  ordered by decreasing eigenvalue.
  The third principal axis $s = e_3/\|e_3\|$ is used as an approximation of the superior–inferior (SI) direction.
  We remove the SI component,
  \[
    x_i^\perp = (x_i - c) - \bigl((x_i-c)^\top s\bigr)s,
  \]
  and perform a second PCA on $\{x_i^\perp\}$ to obtain two orthonormal in-plane axes
  $u,v\in\mathbb{R}^3$.
  Each point is then represented in local planar coordinates
  \[
    (u_i,v_i) = \bigl(u^\top(x_i-c),\, v^\top(x_i-c)\bigr)\in\mathbb{R}^2 .
  \]

  \item \textbf{Planar clustering.}
  On the 2D coordinates $z_i=(u_i,v_i)$ we run $k$-means with $k=2$,
  obtaining a partition $\{1,2\}$ that minimizes within-cluster variance and cluster centroids
  $\mu_1,\mu_2\in\mathbb{R}^2$.
  These centroids are mapped back to world coordinates as
  \[
    \tilde{x}_j = c + \mu_{j,1} u + \mu_{j,2} v,\qquad j\in\{1,2\},
  \]
  and assigned to lateral vs.\ medial using the scanner left–right axis and the known laterality (L/R) of the knee:
  for a left knee the cluster with larger $x$-coordinate is labelled lateral, whereas for a right knee it is the opposite.

  \item \textbf{Consistent partitioning of all masks.}
  The fitted $k$-means model in $(u,v)$-space is then reused for all relevant masks
  (femoral/tibial cartilage and their subchondral bands).
  For any voxel $y$ in these masks we compute its planar coordinates
  $(u_y,v_y)$ as above, predict its cluster label with the same $k$-means model, and map it to the corresponding
  L/M compartment.
  This yields a single, case-specific L/M boundary that is applied consistently across all structures.
\end{enumerate}

This constitutes a \emph{rules + unsupervised} scheme:
PCA and $k$-means adapt to the individual tibial geometry, while simple laterality rules provide anatomical semantics (lateral vs.\ medial).
Unlike template- or atlas-based cartilage partitioning methods that rely on predefined radial transformations and elastic registration
\cite{hafezi2017prediction, panfilov2022deep}, our L/M boundary is obtained directly from the patient’s tibial shape without any template.

\subsection{Regions of interest (ROIs)}
\label{subsec:radiomics_roi}

All radiomics analyses are based on the final, post-processed 3D masks
obtained in Sections~\ref{subsec:nnunet}--\ref{subsec:postproc}. From
the femoral and tibial bone masks and their corresponding cartilage
masks, we derive fixed-thickness subchondral bone bands using the
Euclidean distance transform in millimetres. 
For radiomics, we focus on $t = 10$\,mm bands to align with prior work
on subchondral bone radiomics in knee OA
\cite{fu2025mri,xue2022radiomics}. All cartilage and subchondral
structures are further split into lateral (L) and medial (M) compartments
using the case-specific geometric L/M boundary.

Because the SKM-TEA and OAIZIB-CM models have slightly different label
sets, we map their outputs to a common set of radiomics ROIs and do not
use the meniscal labels from SKM-TEA in the present analysis. This
yields, for each knee, the following primary ROIs:
\begin{itemize}
  \item Femoral cartilage (FC), split into LF and MF.
  \item Tibial cartilage (TC), split into LT and MT.
  \item Patellar cartilage (PC), split into LP and MP.
  \item Femoral subchondral bone band, split into LF\_SUB and MF\_SUB.
  \item Tibial subchondral bone band, split into LT\_SUB and MT\_SUB.
\end{itemize}

These ROIs provide anatomically interpretable subregions covering both
articular cartilage and the immediately underlying subchondral bone,
which are expected to jointly capture OA-related morphological and
textural changes.

\section{Segmentaion Experiments}
\label{sec:segmentation}

\subsection{Environment Settings}
All the code runs within the Ubuntu 22.04.3 LTS on Windows 10 x86\_64 operating system, with an NVIDIA GeForce RTX 4060 Ti GPU, and utilizes the PyTorch framework. The code for LM-CartSeg can be accessed at https://github.com/jukieCheung/LM-CartSeg. The code for nnU-Net can be accessed at https://github.com/MIC-DKFZ/nnU-Net.

\subsection{Datasets and splits}
\label{subsec:datasets}
We train two separate 3D nnU-Net models \cite{isensee2021nnu,isensee2024nnu}
on the training portions of SKM-TEA \cite{desai2022skm} and
OAIZIB-CM \cite{ambellan2019automated,yao2024cartimorph},
respectively. The SKM-TEA training split contains 138 knee MR volumes,
while the OAIZIB-CM training split contains 404 volumes. All splits are
defined case-wise to ensure that no subject contributes scans to both
training and evaluation.


To assess \emph{cross-domain} and \emph{zero-shot} generalisation, we
apply the trained models without any fine-tuning to three additional
cohorts: the SKI-10 challenge dataset \cite{heimann2010segmentation} thah includes 100 knees with expert ground truth, an official test set of OAIZIB-CM cohort
with 103 knees, and a private Po-OA dataset comprising 185 clinical
knee MR scans. SKI-10 and the held-out OAIZIB-CM cohort are used to
evaluate segmentation performance in genuinely out-of-distribution
settings, whereas Po-OA—where voxelwise ground truth is not available—is
primarily used to study the clinical translation of downstream volume and
radiomics analyses under zero-shot segmentation.

\subsection{Preprocessing}
In geometry, we resample all images to a common voxel spacing is applied to (256, 256, 160), where x=coronal, y=axial, z=sagittal, only when needed; all geometric operations (distance thresholds and band thickness) are defined in \emph{millimeters} to decouple them from resampling.




\subsection{Evaluation protocol}
\label{subsec:evaluation}

\paragraph{Segmentation accuracy}
On OAIZIB-CM and SKI-10 we evaluate the automatic segmentations using
standard overlap and surface metrics: Dice similarity coefficient (DSC),
95th-percentile Hausdorff distance (HD95), and average symmetric surface
distance (ASSD).\cite{dice1945measures,heimann2009comparison}
For a given structure with ground-truth label set $G$ and predicted
label set $P$, the DSC is
\begin{equation}
  \mathrm{DSC}(G,P)
  = \frac{2\,|G \cap P|}{|G| + |P|}.
\end{equation}
Surface-based distances are computed on the corresponding object
surfaces $\partial G$ and $\partial P$. Let $d(x,\partial P)$ denote the
minimal Euclidean distance from a point $x$ on $\partial G$ to
$\partial P$ (and analogously $d(y,\partial G)$ for $y \in \partial P$).
The \emph{directed} surface distance distributions are
\begin{equation}
  D_{G\to P} = \{ d(x,\partial P) : x \in \partial G \},\quad
  D_{P\to G} = \{ d(y,\partial G) : y \in \partial P \}.
\end{equation}
We then define
\begin{align}
  \mathrm{HD95}(G,P)
  &= \max\!\bigl\{
      \operatorname{percentile}_{95}(D_{G\to P}),
      \operatorname{percentile}_{95}(D_{P\to G})
    \bigr\},\\
  \mathrm{ASSD}(G,P)
  &= \frac{1}{|\partial G| + |\partial P|}
     \left(
       \sum_{x \in \partial G} d(x,\partial P)
       + \sum_{y \in \partial P} d(y,\partial G)
     \right).
\end{align}
These metrics are computed per structure (femur, tibia, patella, femoral cartilage, tibial cartilage, patellar cartilage, and 10\,mm subchondral bands) using a common evaluation script based on surface-distance calculations, and summarised as case-wise macro averages over labels:
\begin{equation}
  \overline{\mathrm{DSC}}^{(i)}
  = \frac{1}{L} \sum_{\ell=1}^{L}
    \mathrm{DSC}\bigl(G^{(i)}_\ell, P^{(i)}_\ell\bigr),
\end{equation}
for case $i$ and $L$ evaluated labels.

\paragraph{L/M partition accuracy}
To assess the quality of the lateral/medial (L/M) split of femoral and
tibial cartilage, we compare two strategies under identical ground
truth: (A) a direct nnU-Net trained to predict L/M compartments as the \textit{pred}, and
(B) our two-stage approach that first segments whole cartilage and then
applies a geometric L/M partition as the \textit{GT}. Let $\Omega$ denote the set of voxels
in the tibial cartilage region of interest, and let
$y^{\text{GT}}(v) \in \{\mathrm{LT},\mathrm{MT}\}$ and
$y^{\text{pred}}(v) \in \{\mathrm{LT},\mathrm{MT}\}$ denote the
ground-truth and predicted L/M labels for voxel $v \in \Omega$. The
overall L/M accuracy is
\begin{equation}
  \mathrm{ACC}_{\mathrm{LM}}
  = \frac{1}{|\Omega|}
    \sum_{v \in \Omega}
    \mathbb{I}\bigl( y^{\text{GT}}(v) = y^{\text{pred}}(v) \bigr),
\end{equation}
and class-wise precision/recall are derived from the usual confusion
counts for LT and MT.

To quantify mixing along the inter-compartment boundary, we define a
\emph{boundary confusion} rate. Let $\Gamma$ be a one-voxel-wide band
around the LT/MT interface, such as the union of dilated LT and MT
surfaces. The symmetric boundary confusion is
\begin{equation}
  \mathrm{BC}
  = \frac{1}{|\Gamma|}
    \sum_{v \in \Gamma}
    \mathbb{I}\bigl( y^{\text{GT}}(v) \ne y^{\text{pred}}(v) \bigr),
\end{equation}
and directional confusion (LT$\to$MT, MT$\to$LT) is obtained by
restricting the sum to voxels with $y^{\text{GT}}(v)=\mathrm{LT}$ or
$=\mathrm{MT}$, respectively. Alongside per-compartment DSC and HD95 for
LF/MF and LT/MT, these measures quantify how much the predicted L/M
labels ``leak'' across the anatomical boundary when comparing direct
L/M prediction to our geometric partition.

\paragraph{Anatomical quality control (QC)}
Beyond voxelwise overlap, we perform automatic anatomical QC on all
predicted $10$\,mm cartilage and subchondral bone bands, inspired by
prior work on segmentation quality control
\cite{heimann2009comparison,fournel2021medical, guo2024automatedthickness, specktor2025segqc}.
For a region of interest $R$ with voxel set $\mathcal{V}_R$ and voxel
volume $\Delta v$, the anatomical volume is
\begin{equation}
  V_R = |\mathcal{V}_R| \,\Delta v.
\end{equation}
Let $A_R$ denote the surface area of $R$ obtained from a triangulated
surface mesh, such as marching cubes. A simple global thickness estimate
is then
\begin{equation}
  \bar{T}_R = \frac{2 V_R}{A_R},
\end{equation}
and, when needed, local thickness distributions are summarised using
percentiles of a 3D Euclidean distance transform inside $R$.

For lateral--medial comparisons, we compute thickness ratios and
differences, for example for tibial cartilage
\begin{equation}
  r_{\mathrm{TIB}}
  = \frac{\bar{T}_{\mathrm{MT}}}{\bar{T}_{\mathrm{LT}}},\qquad
  \Delta_{\mathrm{TIB}}
  = \bar{T}_{\mathrm{MT}} - \bar{T}_{\mathrm{LT}},
\end{equation}
and analogously for femoral cartilage. At the cohort level, we inspect
the distributions of $V_R$, $\bar{T}_R$, $r_{\mathrm{TIB}}$ and
$r_{\mathrm{FEM}}$ (means, standard deviations and coefficients of
variation) and perform paired Wilcoxon tests for LT--MT and LF--MF
comparisons. Cases with implausible morphometry, like extremely small
$V_R$, extreme $\bar{T}_R$ or thickness ratios far from 1, are flagged
as outliers for visual review. These QC checks ensure that the
automatically generated ROIs are not only accurate with respect to the
reference segmentations, but also anatomically plausible across
datasets.

\section{Radiomics Experiments}
\label{sec:radiomics}

\subsection{Datasets and labels}
\label{subsec:radiomics_datasets}

Radiomic analysis was performed on the held-out official test set of the OAIZIB-CM with 103 knees \cite{ambellan2019automated,yao2024cartimorph}
and on an independent in-house knee MRI cohort Po-OA dataset with 185 knees.
For OAIZIB-CM, Kellgren--Lawrence (KL) grades are provided as part of the
metadata, enabling supervised learning with osteoarthritis (OA) labels.
Following common practice, we defined a binary outcome
\textit{radiographic OA} as
\[
\text{OA} =
\begin{cases}
1, & \text{if } \mathrm{KL} \ge 2, \\
0, & \text{if } \mathrm{KL} \le 1.
\end{cases}
\]

For Po-OA's dataset, standard KL grading was not available. Instead, we
reviewed the radiology reports and used the presence of phrases such as
\emph{``osteoarthritic changes''}, \emph{``degenerative changes''},
\emph{``marginal osteophytes''}, or \emph{``patellofemoral osteoarthritic
changes''} as an imaging-report--defined OA indicator. Subjects whose
reports contained at least one of these phrases were labelled as OA; all
others were labelled as non-OA. This definition is not a gold standard,
but is used as a pragmatic, imaging-derived phenotype for exploratory
analysis and for assessing the behaviour of our automated pipeline on
non-OAI data.

Unless otherwise stated, quantitative classification experiments and
model selection were carried out on the OAIZIB-CM test set, whereas Po-OA's
dataset was used to qualitatively assess the external behaviour of
selected signatures and their feature distributions.

\subsection{Feature extraction}
\label{subsec:radiomics_features}

All radiomic features are computed on the original MR images used in the
segmentation pipeline, including 3D DESS-like \cite{ambellan2019automated,yao2024cartimorph, heimann2010segmentation}, T\textsubscript{2}-weighted,
fat-suppressed sequences), rather than on quantitative T\textsubscript{2}
maps. This differs from some prior knee OA radiomics work based on
T\textsubscript{2} mapping \cite{peuna2018variable, kim2021fat}, and may partly explain differences in absolute
feature values and discriminative performance.

Within each ROI, we extract a standard panel of IBSI-compliant
handcrafted radiomic features using an open-source library: PyRadiomics \cite{van2017computational}, to:

\begin{itemize}
  \item \textbf{Gray-level discretisation:} intensities are discretised
        using a fixed bin width: 5 gray levels, harmonised to
        the dynamic range of the sequence \cite{duron2019gray}.
  \item \textbf{First-order statistics:} histograms and intensity
        summary measures (mean, median, percentiles, skewness,
        kurtosis, energy, etc.).
  \item \textbf{Texture features:} gray-level co-occurrence matrix
        (GLCM), run-length matrix (GLRLM), size-zone matrix (GLSZM),
        neighbourhood gray-tone difference matrix (NGTDM), and gray-level
        dependence matrix (GLDM) features, computed on the
        discretised images.
  \item \textbf{Filtered features:} the above first-order and texture
        features are also computed on filtered versions of the images,
        including logarithmic (LoG) filters and multi-scale
        wavelet decompositions, thereby capturing multi-scale texture
        patterns.
\end{itemize}

Shape features are extracted for completeness but, as discussed below,
they are excluded from the primary radiomics signatures due to their low
robustness under segmentation perturbations and redundancy with simpler
morphological measures that is the volume. Unless otherwise noted, all
classification models in Section~\ref{subsec:radiomics_classification}
are built on non-shape radiomic features.

\subsection{Feature selection and classification}
\label{subsec:radiomics_classification}

To distinguish OA (KL~$\ge 2$) from non-OA knees in the OAIZIB-CM test
set, we explore several supervised classification pipelines that combine
feature selection with regularised classifiers. All models are evaluated
using stratified $k$-fold cross-validation (typically $k=5$), and
performance is reported as the mean area under the ROC curve (AUC) and
accuracy across folds.

\paragraph{Statistical feature selection pipelines}
We consider a statistical feature selection schemes motivated by
existing radiomics studies:

\begin{itemize}

  \item \textbf{Xue-style pipeline. \cite{xue2022radiomics}} As an alternative, we first apply
        LASSO logistic regression \cite{tibshirani1996regression} to the full non-shape feature set to
        obtain a preliminary sparse subset. For these selected features,
        we then perform Mann--Whitney U tests to retain only those with
        significant group differences and optionally remove highly
        collinear features using a correlation threshold, where Pearson
        $|r| > 0.9$. 
\end{itemize}

These baselines reflect conventional radiomics practice where feature
selection is driven purely by predictive performance and statistical
significance, without explicit robustness considerations.


\paragraph{Classifiers and evaluation}
For each selected feature set, we train the L1-regularised logistic regression (LASSO), which simultaneously performs feature selection.

Hyperparameters are set as default. For each model and feature
set, we record 5 folds cross-validated AUC and accuracy. 

Together, these experiments position our work within the broader
radiomics framework: starting from fully automatic, anatomically guided
ROIs, we systematically examine the interplay between segmentation
robustness, feature selection strategy, and downstream classification
performance for MRI-based knee OA assessment.


\section{Segmentation Results}

\subsection{Training and in-domain validation}

The two 3D nnU-Net models converged stably on their respective training
cohorts. On SKM-TEA, the mean validation Dice similarity coefficient
(DSC) across all labels was $0.867$, while on OAIZIB-CM it reached
$0.916$, indicating accurate multi-structure segmentation on both
datasets.

\subsection{Segmentation accuracy on the OAIZIB-CM test set}

\begin{table}[t]
\centering
\caption{Macro-averaged segmentation metrics on the OAIZIB-CM test set
(103 knees) before and after geometric post-processing. $\Delta$ denotes
clear\_seg $-$ seg.}
\label{tab:oai_macro}
\begin{tabular}{lcccc}
\hline
Metric & seg & clear\_seg & $\Delta$ & Wilcoxon $p$ \\
\hline
DSC $\uparrow$          & 0.892 & 0.907 & +0.015 & $4.5\times 10^{-10}$ \\
ASSD [mm] $\downarrow$    & 2.63  & 0.36  & $-2.26$ & $1.3\times 10^{-18}$ \\
HD95 [mm] $\downarrow$    & 25.16 & 3.35  & $-21.81$ & $2.4\times 10^{-18}$ \\
\hline
\end{tabular}
\end{table}

\begin{figure}
\centering
\includegraphics[width=\linewidth]{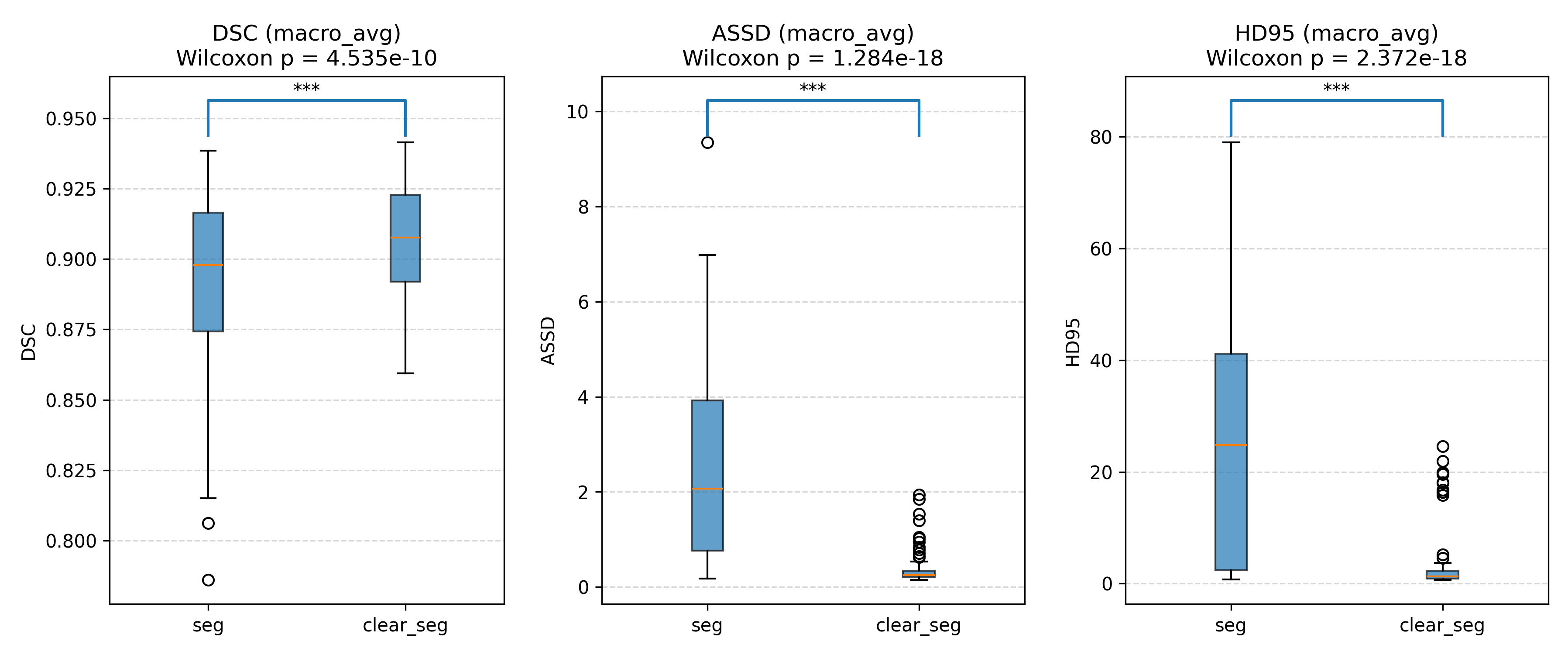}
\caption{Macro-averaged DSC, ASSD and HD95 on the OAIZIB-CM test set,
comparing raw nnU-Net segmentations (seg) with post-processed masks
(clear\_seg). Brackets indicate paired Wilcoxon $p$-values.}
\label{fig:oai_boxplots}
\end{figure}

On the official OAIZIB-CM test cohort (103 knees), we compared the raw
nnU-Net predictions with the geometrically cleaned LM-CartSeg outputs,
using case-wise macro-averaged DSC, average symmetric surface distance
(ASSD) and 95th-percentile Hausdorff distance (HD95) over the four
evaluated structures. As summarised in Table~\ref{tab:oai_macro} and Figure~\ref{fig:oai_boxplots}, the
post-processing step substantially improved surface agreement: mean ASSD
decreased from $2.63$\,mm to $0.36$\,mm and mean HD95 from
$25.2$\,mm to $3.35$\,mm, while the mean DSC increased from $0.892$ to
$0.907$. Paired Wilcoxon signed-rank tests on the
macro-averaged metrics confirmed that these improvements were highly
significant (DSC $p = 4.5\times 10^{-10}$, HD95
$p = 2.4\times 10^{-18}$, ASSD $p = 1.3\times 10^{-18}$).

Label-wise analyses showed the same pattern for three out of four
structures, with consistent and significant gains in all metrics; for the
remaining structure, DSC was statistically unchanged (Wilcoxon
$p = 0.12$) but HD95 and ASSD still improved markedly after
post-processing. Overall, these results demonstrate
that the geometric cleaning and instance-merging rules lead to
systematically smoother and more anatomically plausible segmentations
without sacrificing overlap accuracy.

\subsection{Zero-shot performance on SKI-10}

We further evaluated cross-domain generalisation on the SKI-10 challenge
cohort (100 knees), where neither images nor labels were seen during
training. In this strictly zero-shot setting, the raw nnU-Net
predictions already achieved a macro-averaged DSC of $0.796$, ASSD of
$1.07$\,mm and HD95 of $7.59$\,mm across the four evaluated structures.

\begin{table}[t]
\centering
\caption{Macro-averaged zero-shot segmentation performance on the SKI-10
dataset (100 knees). Values are means across cases; $\Delta$ denotes
clear\_seg $-$ seg.}
\label{tab:ski_macro}
\begin{tabular}{lcccc}
\hline
Metric      & seg   & clear\_seg & $\Delta$ & Wilcoxon $p$ \\
\hline
DSC $\uparrow$        & 0.796 & 0.797      & +0.001   & $3.3\times 10^{-5}$ \\
ASSD [mm] $\downarrow$  & 1.07  & 0.94       & $-0.13$  & $1.3\times 10^{-5}$ \\
HD95 [mm] $\downarrow$  & 7.59  & 6.13       & $-1.46$  & $1.8\times 10^{-1}$ \\
\hline
\end{tabular}
\end{table}

\begin{figure}
\centering
\includegraphics[width=\linewidth]{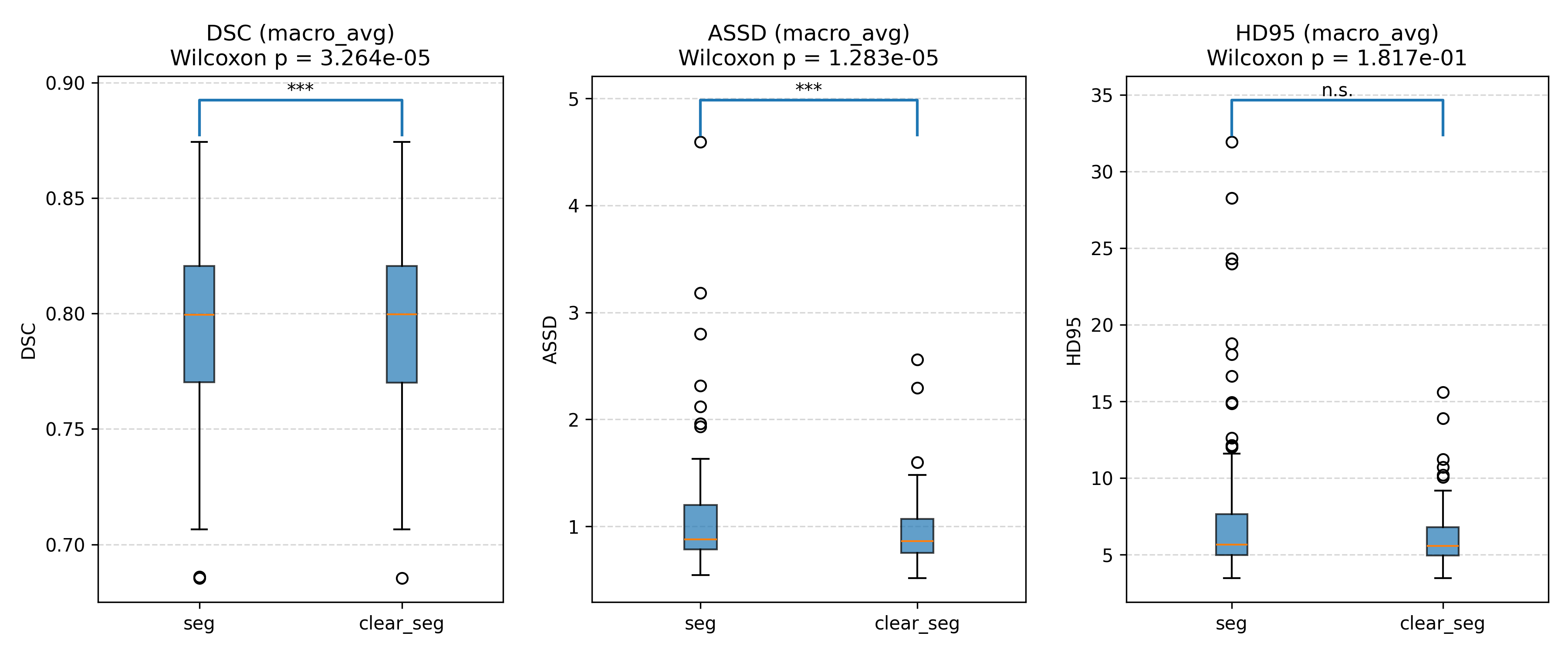}
\caption{Zero-shot macro-averaged DSC, ASSD and HD95 on the SKI-10
dataset, comparing seg vs.\ clear\_seg. Brackets show paired Wilcoxon
$p$-values; HD95 differences are not significant.}
\label{fig:ski_boxplots}
\end{figure}

After applying the proposed geometric cleaning and instance-merging
rules, macro-averaged ASSD improved from $1.07$\,mm to $0.94$\,mm and
HD95 from $7.59$\,mm to $6.13$\,mm, while DSC increased slightly from
$0.796$ to $0.797$.
Numerical results are summarised in Table~\ref{tab:ski_macro}, with the corresponding distributions shown in Figure~\ref{fig:ski_boxplots}.
Paired Wilcoxon tests on the macro-averaged metrics showed significant
gains in DSC ($p = 3.3\times 10^{-5}$) and ASSD
($p = 1.3\times 10^{-5}$), whereas the HD95 reduction did not reach
statistical significance ($p = 0.18$).

Label-wise analysis revealed that two large structures reached DSC
values above $0.96$ in the zero-shot setting, while the two smaller
structures remained more challenging (DSC $\approx 0.56$–$0.68$). For
three of the four labels, post-processing significantly improved at
least one of DSC, ASSD or HD95, and for the remaining label the overlap
was preserved while surface distances still decreased.

Overall, these results indicate that the proposed geometric rules
consistently enhance surface accuracy without degrading overlap, even on
a dataset that is markedly different from the training domain.

\subsection{Accuracy of the learned L/M tibial partition}
\label{subsec:lm_accuracy}

\begin{table}[t]
\centering
\caption{Voxel-wise LT/MT classification performance of the direct L/M
nnU-Net relative to the geometric L/M partition on the tibial plateau.
Values are mean $\pm$~SD across knees, reported as percentages.}
\label{tab:lm_partition_datasets}
\scalebox{0.7}{
\begin{tabular}{lccc}
\hline
Metric [\%]      & OAIZIB-CM (in-domain) & SKI-10 (zero-shot) & Po-OA (clinical) \\
\hline
Accuracy (ACC)        & 99.9 $\pm$ 0.4  & 92.1 $\pm$ 15.2 & 93.6 $\pm$ 21.9 \\
Balanced accuracy     & 100.0 $\pm$ 0.4 & 92.0 $\pm$ 14.6 & 93.6 $\pm$ 21.8 \\
LT$\rightarrow$MT confusion & 0.0 $\pm$ 0.2   & 9.5 $\pm$ 22.1  & 0.3 $\pm$ 2.0  \\
MT$\rightarrow$LT confusion & 0.1 $\pm$ 0.6   & 4.8 $\pm$ 14.3  & 1.5 $\pm$ 5.4  \\
Symmetric confusion (CONF\_sym) & 0.0 $\pm$ 0.4 & 7.2 $\pm$ 13.8 & 0.9 $\pm$ 2.8  \\
\hline
\end{tabular}
}
\end{table}

We compared our rule-based two-stage L/M partition against a direct
nnU-Net trained to predict lateral tibial (LT) and medial tibial (MT)
cartilage as separate labels. In the two-stage scheme, nnU-Net segments
the whole tibial cartilage and the L/M split is imposed afterwards by
the geometric rule from Section~\ref{subsec:postproc}; by construction,
this partition cannot swap sides and serves as a deterministic reference as the \textit{GT}). The direct L/M nnU-Net as the \textit{pred} is always
evaluated against this GT.

\begin{figure}
\centering
\includegraphics[width=\linewidth]{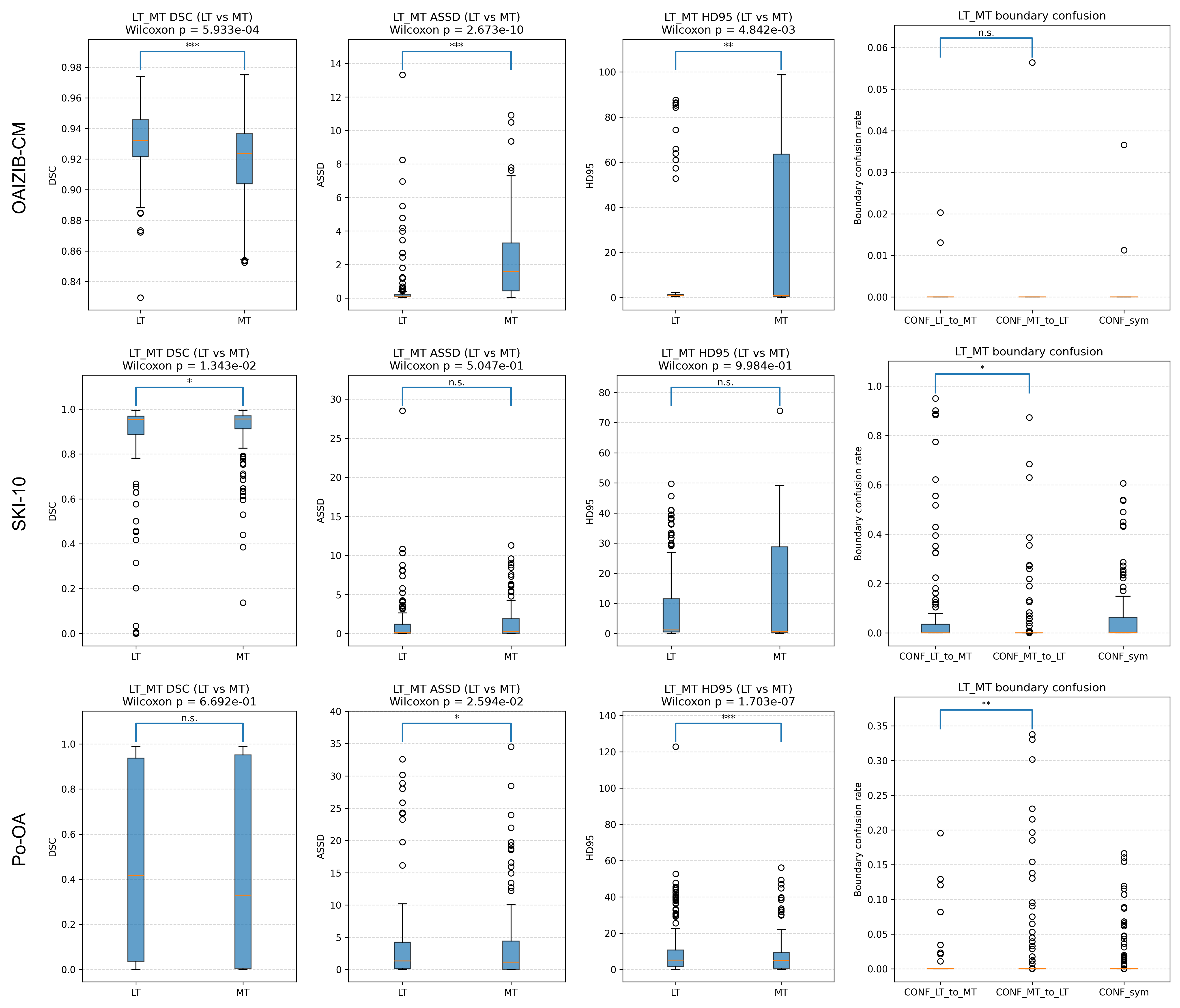}
\caption{Accuracy of the learned tibial L/M partition across datasets.
Rows (top to bottom) show OAIZIB-CM (in-domain), SKI-10 (zero-shot) and
Po-OA (clinical) results for the direct L/M nnU-Net evaluated against the
geometric L/M partition. Columns (left to right) display per-side DSC,
ASSD and HD95 for LT vs.\ MT, and voxel-wise L/M boundary confusion
(CONF\_LT\_to\_MT, CONF\_MT\_to\_LT, CONF\_sym). Brackets indicate paired
Wilcoxon $p$-values. The learned L/M labels closely match the geometric
partition in-domain, but become unstable and asymmetric under domain
shift on SKI-10 and Po-OA.}
\label{fig:ltmt_eval}
\end{figure}

On the in-domain OAIZIB-CM test set, the direct L/M network closely
reproduces the geometric partition, as shown in Figure \ref{fig:ltmt_eval}. DSC, ASSD and HD95 are high and tightly
distributed for both LT and MT, with LT being slightly easier (all
Wilcoxon $p<10^{-2}$). Voxel-wise L/M classification on the tibial
plateau confirms this: the mean accuracy is $99.95\%\pm0.42\%$ and the
balanced accuracy $99.95\%\pm0.39\%$, with per-side precision/recall all
above $99.9\%$. The mean symmetric boundary confusion rate is only
$0.05\%\pm0.38\%$, and there is no significant asymmetry between
LT$\rightarrow$MT and MT$\rightarrow$LT errors, as shown in Table~\ref{tab:lm_partition_datasets}. Thus, within the training domain, a
network can effectively learn the same L/M split as the geometric rule.

In contrast, the zero-shot experiments on SKI-10 and Po-OA show that the
learned L/M labels are much less robust under domain shift. On SKI-10,
many knees still achieve high DSC with respect to the geometric GT, but
there are several clear outliers with DSC near zero for one compartment,
indicating almost complete side swaps. The corresponding boundary
confusion distributions are heavy-tailed, with some cases exhibiting
substantial fractions of tibial-plateau voxels assigned to the wrong
side. On the clinical Po-OA cohort the situation is even more unstable:
per-side DSC distributions are broad with many low values, ASSD and HD95
are highly variable, and the symmetric confusion rate can reach
$>0.3$ in difficult scans. Moreover, LT$\rightarrow$MT errors are
consistently larger than MT$\rightarrow$LT, revealing a systematic
directional bias.

Overall, these results suggest that while a direct L/M nnU-Net can mimic
the geometric partition when train and test domains match, its L/M
labels become unreliable and biased on out-of-distribution data. In
contrast, the proposed geometric L/M rule is deterministic,
annotation-free and domain-agnostic, and is therefore adopted as the
default partition strategy in our downstream radiomics pipeline.

\subsection{Anatomical quality control (QC)}
\label{subsec:qc}

\begin{figure}
\centering
\includegraphics[width=\linewidth]{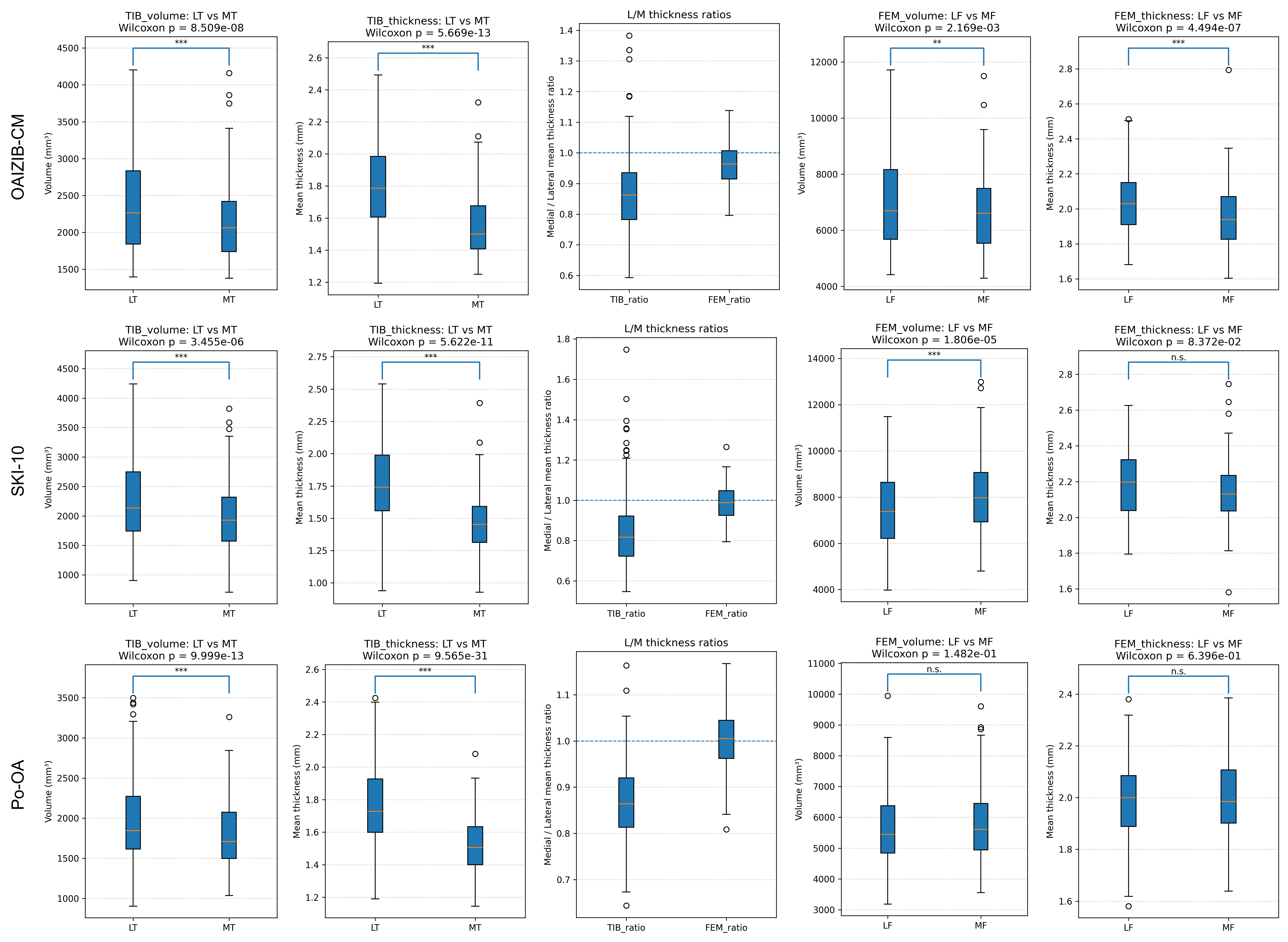}
\caption{Anatomical QC of $10$\,mm cartilage bands across datasets.
Rows (top to bottom) show OAIZIB-CM, SKI-10 and Po-OA; columns (left to
right) show tibial cartilage volume (LT vs.\ MT), tibial mean thickness,
medial/lateral tibial and femoral thickness ratios, femoral cartilage
volume (LF vs.\ MF) and femoral mean thickness. Boxplots are case-wise
distributions and brackets indicate paired Wilcoxon $p$-values; the
dashed horizontal line marks a ratio of 1 (no L/M asymmetry).}
\label{fig:qc_lm}
\end{figure}

To verify that the automatically derived ROIs preserve plausible knee
anatomy, we performed simple QC analyses on the $10$\,mm tibial and
femoral cartilage bands across all three cohorts. For each knee we
computed the volume and mean thickness of the lateral and medial tibial
cartilage (LT, MT) and femoral cartilage (LF, MF), as well as the
medial-to-lateral thickness ratios
(TIB\_ratio, FEM\_ratio). LT--MT and LF--MF pairs were compared with
paired Wilcoxon tests as shown in Figure~\ref{fig:qc_lm}, and summary statistics are
reported in Table~\ref{tab:qc_lm}.

\begin{table}[t]
\centering
\caption{Anatomical QC: lateral/medial tibial and femoral cartilage
thickness asymmetry across datasets. Values are mean$\pm$SD across
knees; ratios are medial/lateral, and $\Delta t$ is medial minus
lateral thickness in mm (negative values indicate thinner medial
cartilage).}
\label{tab:qc_lm}
\scalebox{0.8}{
\begin{tabular}{lcccc}
\hline
Dataset & Tibia ratio & Tibia $\Delta t$ (mm) & Femur ratio & Femur $\Delta t$ (mm) \\
\hline
OAI\textendash ZIB\textendash CM (n=103) & 0.88 $\pm$ 0.14 & $-0.24 \pm 0.25$ & 0.96 $\pm$ 0.07 & $-0.08 \pm 0.15$ \\
SKI-10 (n=100)                           & 0.86 $\pm$ 0.21 & $-0.30 \pm 0.36$ & 0.99 $\pm$ 0.09 & $-0.04 \pm 0.20$ \\
Po-OA (n=185)                            & 0.87 $\pm$ 0.08 & $-0.25 \pm 0.17$ & 1.00 $\pm$ 0.06 & $0.00 \pm 0.12$ \\
\hline
\end{tabular}
}
\end{table}

Across OAIZIB-CM, SKI-10 and Po-OA, tibial cartilage consistently showed
the expected asymmetry: medial tibial cartilage was smaller and thinner
than lateral cartilage. Mean LT volumes exceeded MT volumes in every
dataset, such as \ $2379$ vs.\ $2133$\,mm$^3$ in OAI; $2288$ vs.\
$2004$\,mm$^3$ in SKI-10; $1967$ vs.\ $1782$\,mm$^3$ in Po-OA, and mean
medial--lateral thickness differences were around $-0.25$ to
$-0.30$\,mm, TIB\_ratio $\approx 0.86$--$0.88$ in all three cohorts,
all $p\ll 0.001$. These consistent patterns across domains argue
against systematic L/M swapping or gross geometric artefacts in the
tibial compartments.

Femoral cartilage displayed a more balanced L/M morphology, again in
line with prior anatomical observations. Medial and lateral femoral
volumes were similar that LF and MF within $\sim10\%$ in all cohorts, and
mean thicknesses differed only slightly, with FEM\_ratio close to~1. The LF--MF thickness
difference was significant only in OAI, and non-significant in SKI-10 and
Po-OA, indicating that our pipeline does not impose an artificial L/M
bias on femoral cartilage. Overall, the cross-dataset agreement of these
simple anatomical signatures supports the validity of the automated
segmentations and L/M partitions used for downstream radiomics.

\section{Radiomics Results}

\subsection{Radiomic similarity between compartments}

\begin{figure}
\centering
\includegraphics[width=\linewidth]{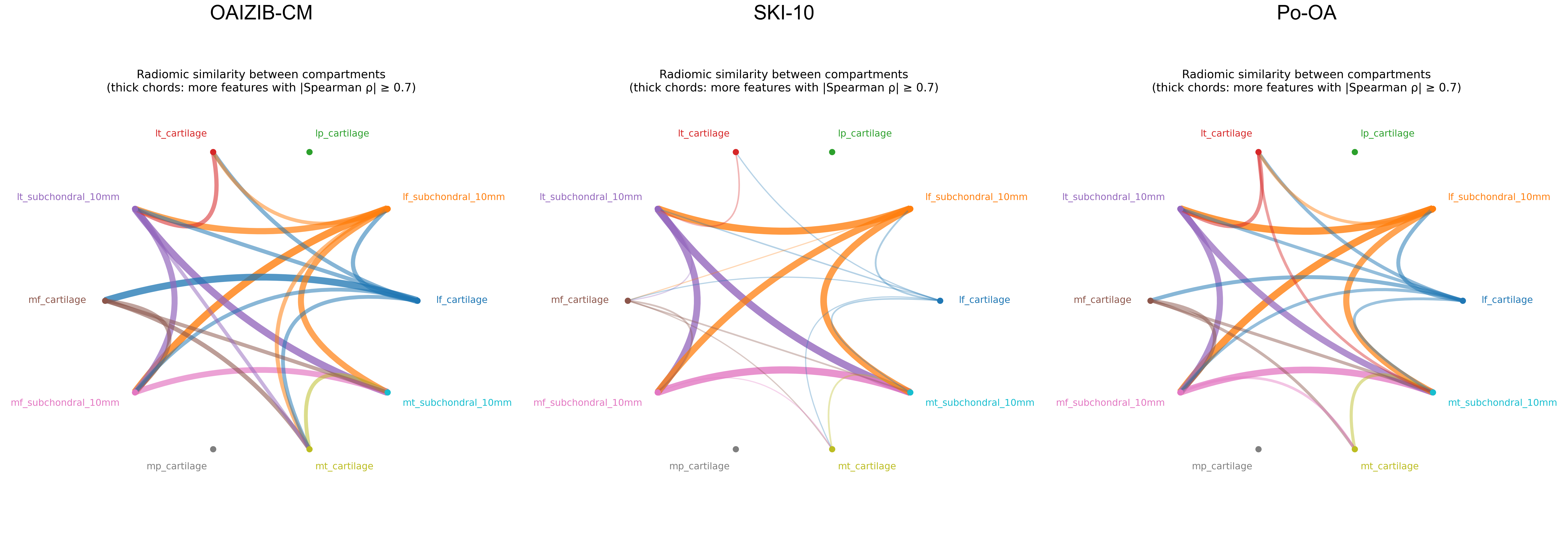}
\caption{Radiomic similarity between knee compartments across datasets.
Chord diagrams show pairwise Spearman correlations \cite{rovetta2020raiders} between radiomic
features of 10 ROIs for OAIZIB-CM (left), SKI-10 (middle) and Po-OA
(right). Nodes correspond to cartilage and 10\,mm subchondral bone
bands; thicker chords indicate more features with $|\rho|\ge 0.7$
between a given pair of compartments.}
\label{fig:radiomics_chord}
\end{figure}

For each of the ten ROIs (LF, MF, LT, MT, LP, MP cartilage and the corresponding 10\,mm subchondral bone bands) we extracted 93 non-shape radiomic features, including 18 first order, 24 GLCM, 14 GLDM, 16 GLRLM, 16 GLSZM, 5 NGTDM, from the original image and four Laplacian of Gaussian (LoG) filtered images, where $\sigma \in \{1, 1.5, 2, 2.5\}$\,mm, yielding $10 \times 93 \times 5 = 4{,}650$ features per dataset.  Within each cohort (OAIZIB-CM, SKI-10, Po-OA) we computed
Spearman correlations between all features for every pair of ROIs and, for each
pair, counted the number of features with $|\rho|\ge0.7$.  These counts are
visualised as chord diagrams in Figure ~\ref{fig:radiomics_chord}, where thicker
chords indicate more strongly correlated features.

Across all three datasets we observed highly consistent patterns.  Subchondral
bone bands showed the strongest radiomic coupling: medial–lateral pairs within the same bone and tibial–femoral subchondral pairs shared the largest numbers of strongly correlated features. 

For instance, in LF\_SUB–MF\_SUB, OAIZIB-CM, SKI-10, and Po-OA up to 29, 90, and 39 features respectively.  Cartilage in the same bone also exhibited substantial
similarity (LF–MF, LT–MT), and each cartilage was radiomically similar to its
underlying subchondral band.  In contrast, patellar cartilage (LP/MP) shared
almost no strongly correlated features with tibiofemoral cartilage or
subchondral bone, appearing nearly isolated in all three chord diagrams.  These
patterns suggest that tibial and femoral cartilage–bone units form tightly
coupled radiomic compartments, whereas the patellofemoral compartment is
radiomically distinct.

\subsection{Relationship between ROI size and radiomic features}

\begin{figure}
\centering
\includegraphics[width=\linewidth]{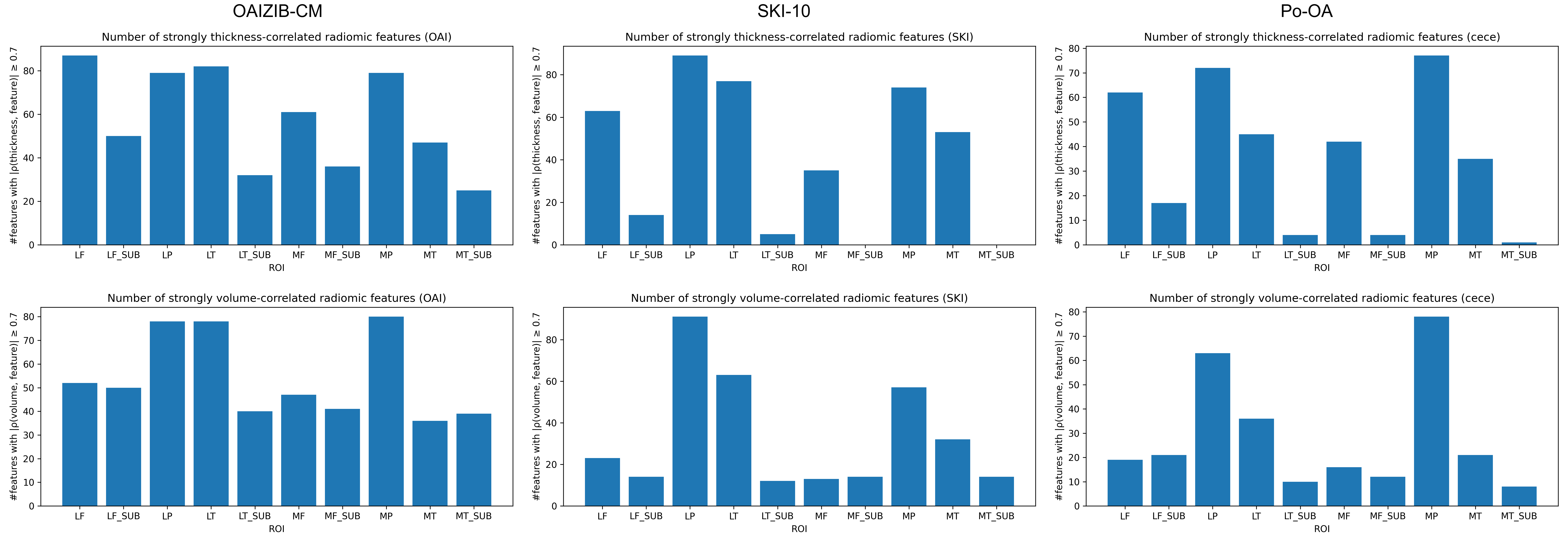}
\caption{Number of radiomic features strongly correlated with ROI size.
For each ROI and dataset (OAIZIB-CM, SKI-10, Po-OA), bar plots show how
many of the 465 radiomic features have Spearman correlation
$|\rho|\ge 0.7$ with mean thickness (top row) or volume (bottom row).
Cartilage ROIs, especially patellar and tibial cartilage, exhibit more
size-correlated features than subchondral bone bands, and thickness
correlations generally exceed volume correlations.}
\label{fig:vol_thick_corr}
\end{figure}

To assess how many radiomic features are essentially proxies for ROI
size, we computed Spearman correlations between each feature and the
corresponding ROI's mean thickness and volume, separately for each
dataset. For every ROI (465 features) we counted the number of features
with strong correlation $|\rho|\ge 0.7$ with thickness or volume in Figure ~\ref{fig:vol_thick_corr}).

Across cohorts, only a minority of features showed such strong
size-dependence. On average, $6$--$12\%$ of features per ROI were
strongly correlated with either thickness or volume, with slightly more
features linked to thickness than to volume (mean proportion of
thickness-correlated features $0.12$ vs.\ $0.08$ for volume across all
ROIs and datasets). Cartilage ROIs were noticeably more size-dependent
than subchondral bands: in OAIZIB-CM, cartilage had
$\sim13$--$16\%$ of features strongly correlated with thickness or
volume, whereas the subchondral bands had only $7$--$9\%$; in SKI-10 and
Po-OA the subchondral proportions dropped below $3\%$.

The patellar cartilage (LP/MP) and tibial cartilage (LT/MT) consistently
exhibited the largest numbers of size-correlated features (up to
$\sim90/465$ features in some datasets), indicating that a substantial
subset of their texture descriptors reflects simple geometric
differences. In contrast, femoral and tibial subchondral bands showed
few strongly size-correlated features, suggesting that their radiomic
profiles capture information that is less dominated by global thickness
or volume. Overall, these analyses confirm that most radiomic features
are not trivially explained by ROI size, while also identifying
compartments and feature subsets in which size effects are more
pronounced.

\subsection{OA classification from radiomics versus size-linked features}

We evaluated whether OA classification is mainly driven by simple size
differences (volume/thickness) or by subtler radiomic patterns. For each
dataset (OAIZIB-CM, Po-OA) we considered two feature sets:

(i) a radiomics set obtained from our pipeline after univariate
filtering and LASSO feature selection (\textit{radiomics}); and

(ii) a size-linked set restricted to radiomic features that showed
strong correlation with ROI volume or thickness (|Spearman $\rho|\ge
0.7$; \textit{size-linked}). For both sets we trained four classifiers
: LASSO logistic regression \cite{tibshirani1996regression}, $k$NN \cite{cover1967nearest}, RBF-SVM \cite{scholkopf2002learning}, XGBoost \cite{chen2016xgboost}, in 5-fold
stratified cross-validation and reported AUC, F1, accuracy, precision
and recall in Table~\ref{tab:oa_classification}.

\begin{table}[t]
\centering
\caption{OA vs non-OA classification performance (5-fold CV) on
OAIZIB-CM and Po-OA for radiomics features vs size-linked
(volume/thickness-correlated) features. $N_\text{feat}$ is the number of
features used after LASSO.}
\label{tab:oa_classification}
\scalebox{0.8}{
\begin{tabular}{l l r c c c c c}
\hline
Dataset / set & Model & $N_\text{feat}$ & AUC & F1 & Acc. & Rec. & Prec. \\
\hline
OAIZIB-CM           & lasso    & 14 & 0.906 & 0.832 & 0.792 & 0.812 & 0.852 \\
OAIZIB-CM           & knn     & 14 & 0.868 & 0.827 & 0.772 & 0.859 & 0.797 \\
OAIZIB-CM           & svm\_rbf & 14 & 0.891 & 0.882 & 0.842 & 0.938 & 0.833 \\
OAIZIB-CM           & xgboost & 14 & 0.849 & 0.833 & 0.782 & 0.859 & 0.809 \\
OAIZIB-CM (volume/thickness)  & lasso    &  4 & 0.777 & 0.752 & 0.673 & 0.781 & 0.725 \\
OAIZIB-CM (volume/thickness)  & knn      &  4 & 0.740 & 0.748 & 0.693 & 0.719 & 0.780 \\
OAIZIB-CM (volume/thickness)  & svm\_rbf &  4 & 0.768 & 0.760 & 0.693 & 0.766 & 0.754 \\
OAIZIB-CM (volume/thickness)  & xgboost  &  4 & 0.730 & 0.733 & 0.653 & 0.750 & 0.716 \\
\hline
Po-OA                     & lasso    & 33 & 0.803 & 0.705 & 0.716 & 0.674 & 0.738 \\
Po-OA                     & knn      & 33 & 0.732 & 0.667 & 0.678 & 0.641 & 0.694 \\
Po-OA                     & svm\_rbf & 33 & 0.827 & 0.755 & 0.749 & 0.772 & 0.740 \\
Po-OA                     & xgboost  & 33 & 0.765 & 0.701 & 0.710 & 0.674 & 0.729 \\
Po-OA (volume/thickness)       & lasso    &  5 & 0.729 & 0.685 & 0.694 & 0.663 & 0.709 \\
Po-OA (volume/thickness)       & knn      &  5 & 0.634 & 0.583 & 0.601 & 0.554 & 0.614 \\
Po-OA (volume/thickness)       & svm\_rbf &  5 & 0.682 & 0.644 & 0.656 & 0.620 & 0.671 \\
Po-OA (volume/thickness)       & xgboost  &  5 & 0.613 & 0.551 & 0.563 & 0.533 & 0.570 \\
\hline
\end{tabular}
}
\end{table}

\begin{figure}
\centering
\includegraphics[width=\linewidth]{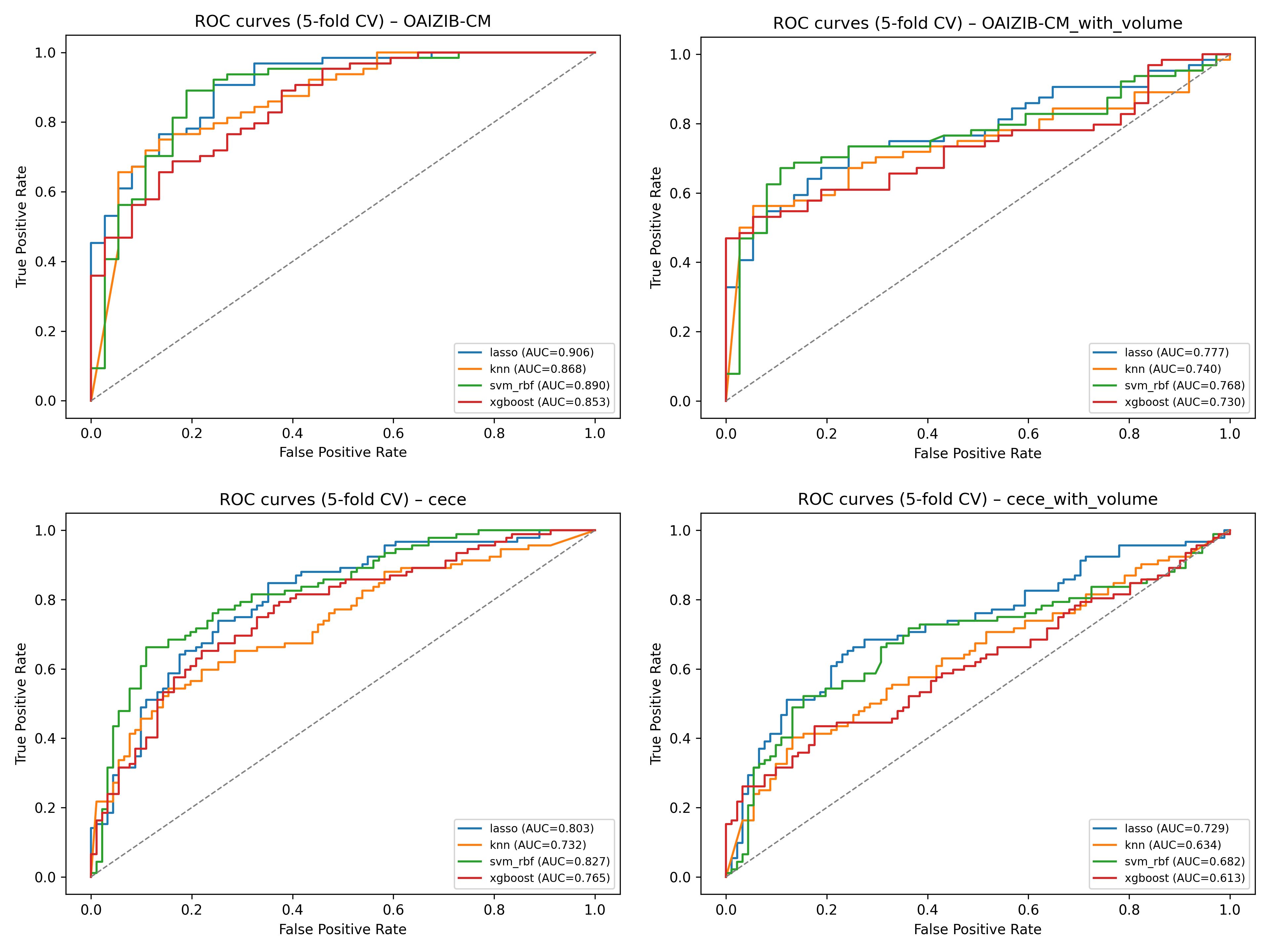}
\caption{OA vs non-OA classification ROC curves (5-fold CV) on
OAIZIB-CM (top) and Po-OA (bottom) for four classifiers (LASSO, $k$NN,
RBF-SVM, XGBoost). Left: models trained on LASSO-selected radiomic
features; right: models trained on size-linked (volume/thickness-
correlated) features only. Legends report the corresponding AUCs. In
both cohorts, radiomic features consistently outperform size-linked
features across all classifiers.}
\label{fig:roc_radiomics}
\end{figure}

On OAIZIB-CM, the radiomics features achieved AUCs between 0.849 and
0.906 across models, with the best result for LASSO (AUC=0.906,
Accuracy=0.79). Restricting the input to size-linked features reduced
AUC to 0.730–0.777, i.e.\ a mean drop of $0.12$–$0.13$ across models.
On the Po-OA cohort, the pattern was similar: radiomics features yielded
AUCs of 0.732–0.827 (best: RBF-SVM, AUC=0.827, Accuracy=0.75), whereas
size-linked features only reached 0.613–0.729 (mean AUC drop
$0.07$–$0.15$). ROC curves in Figure ~\ref{fig:roc_radiomics} illustrate
this consistent performance gap.

\begin{figure}
\centering
\includegraphics[width=\linewidth]{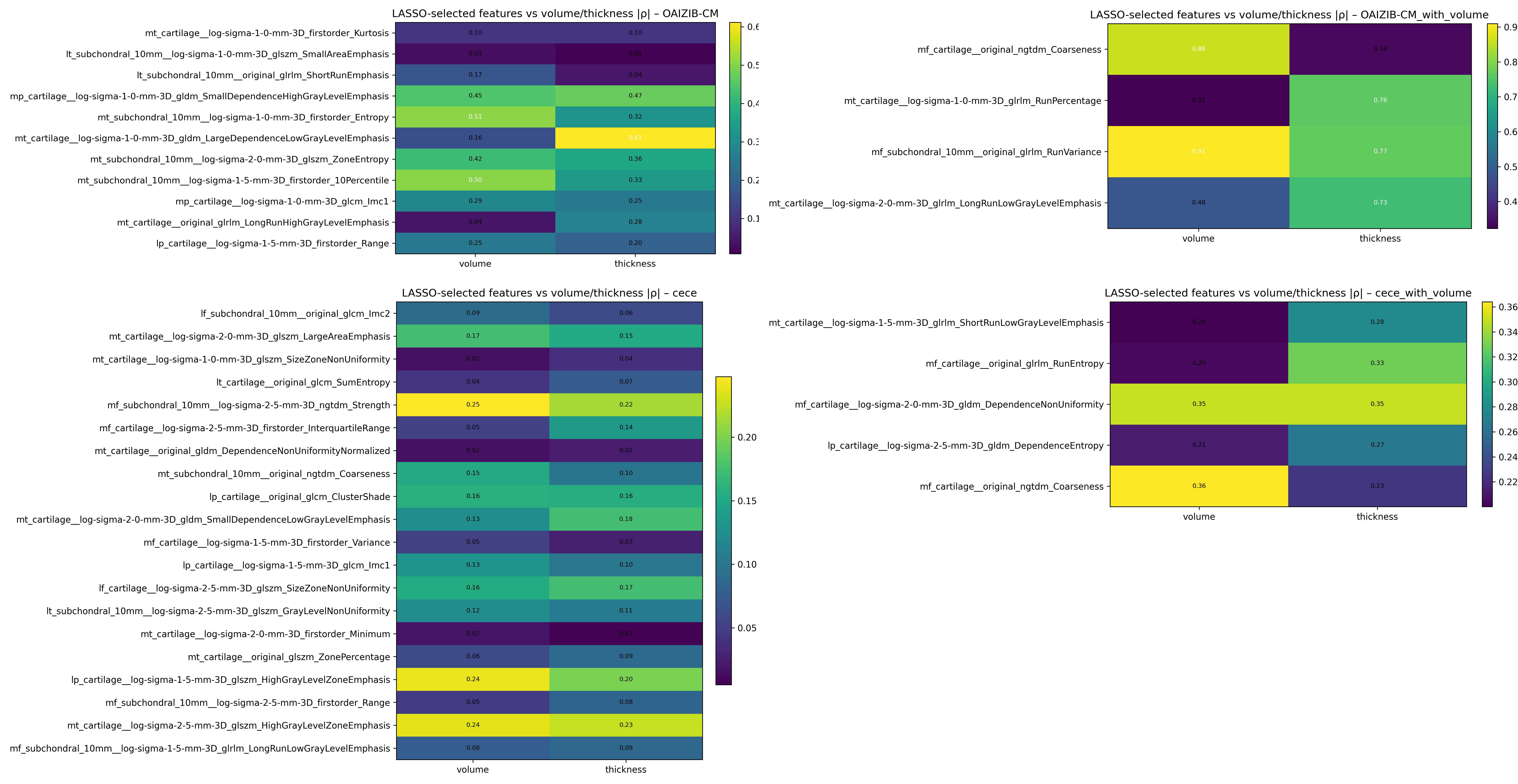}
\caption{Correlation of LASSO-selected radiomic features with ROI size.
Heatmaps show absolute Spearman correlations $|\rho|$ between each
selected feature and its ROI volume/thickness for OAIZIB-CM (top) and
Po-OA (bottom). Left: features selected from the full radiomics sets
(\textit{radiomics}); right: features selected when restricting the
input to volume/thickness-correlated candidates (\textit{with\_volume}).
Radiomics-selected features are only weakly size-dependent, whereas
with\_volume features are, by design, strongly correlated with ROI size.}
\label{fig:lasso_size_corr}
\end{figure}

To further probe size dependence, we examined the Spearman correlations
between the LASSO-selected radiomic features and ROI volume/thickness in Figure ~\ref{fig:lasso_size_corr}. For the radiomics feature sets, most
selected features showed only modest correlation with size
(|$\rho| \lesssim 0.3$–0.4), confirming that LASSO preferentially picks
texture descriptors that are not simple surrogates of volume or
thickness. In contrast, the features selected from the size-linked pools
were, by construction, strongly correlated with size (many
$|\rho|\ge 0.7$), yet they produced clearly inferior OA discrimination.
Overall, these results indicate that our pipeline captures radiomic
information beyond gross morphometric changes, and that this additional
texture signal substantially improves OA vs non-OA classification over
volume/thickness-driven features alone.

\section{Discussion}
In this work we introduced LM-CartSeg, an end-to-end framework that links
state-of-the-art 3D nnU-Net segmentation with rule-based geometric
post-processing, automated L/M partitioning and radiomics analysis of
knee MRI. Across multiple public and clinical cohorts, the pipeline
produced accurate, anatomically plausible ROIs for femoral, tibial and
patellar cartilage and their 10~mm subchondral bone bands, and enabled
systematic investigation of radiomic similarity, size dependence and OA
classification. Compared with prior KOA radiomics studies that
typically target either cartilage or subchondral bone in isolation and
rely on manual delineation of a few compartments \cite{hirvasniemi2021machine, xue2022radiomics, cui2023development, fu2025mri, chen2025integration, angelone2025innovative, lin2023prediction, xie2021radiomics}, our approach offers
fully automatic, multi-compartment ROIs that jointly represent
cartilage--bone units and explicitly encode L/M anatomy.

The segmentation experiments highlight the value of combining nnU-Net
with simple geometric rules. On both the in-domain OAIZIB-CM test set
and the out-of-distribution SKI-10 dataset, post-processing
substantially reduced ASSD and HD95 while preserving or slightly
improving DSC. These gains reflect removal of small spurious components
and smoothing of jagged surfaces, which are known to degrade
surface-based metrics despite high voxelwise overlap. Importantly, we
evaluated not only overlap but also the stability of the learned L/M
partition. A direct L/M nnU-Net could recover the geometric split almost
perfectly on OAIZIB-CM, but under domain shift to SKI-10 and Po-OA it
occasionally swapped sides or produced highly asymmetric boundary
confusion. In contrast, our geometric L/M rule is deterministic,
annotation-free and depends only on tibial shape and scanner left--right
orientation, making it intrinsically robust to changes in intensity,
sequence and OA severity. This suggests that for tasks where anatomical
semantics are largely geometric---such as L/M partitioning---simple
rules can outperform purely data-driven labels in terms of
generalisability.

The anatomical QC analyses further support the validity of the
generated ROIs. Across three datasets, medial tibial cartilage was
consistently smaller and thinner than lateral cartilage, with
medial--lateral thickness ratios around 0.86--0.88 and differences of
$-0.25$ to $-0.30$~mm, in line with known loading patterns and OA
predilection of the medial compartment \cite{shah2019variation, jansen2022knee}. Femoral cartilage exhibited
near-symmetric volumes and thicknesses, with femoral thickness ratios
close to 1 and mostly non-significant L/M differences \cite{giurazza2025femoral}. These consistent
patterns across heterogeneous cohorts argue against systematic L/M
swapping or gross geometric artefacts, and illustrate how simple
morphometric signatures can serve as practical QC indicators in
large-scale automated pipelines.

Our radiomics analyses provide several insights into how cartilage and
subchondral bone contribute to OA-related image phenotypes. First, chord
diagrams showed that radiomic profiles of subchondral bands are strongly
coupled within each bone and between tibial and femoral bands, whereas
patellar cartilage appears radiomically isolated from tibiofemoral
compartments. This supports the notion of bone--cartilage units as
functional entities in OA and is consistent with subchondral bone
radiomics studies that achieved strong discrimination between OA and
non-OA knees \cite{hirvasniemi2021machine, xue2022radiomics, fu2025mri}. Second, only a minority of features---typically 6--12\%
per ROI---were strongly correlated with volume or thickness, and
correlations were weaker in subchondral bands than in cartilage. Thus,
most radiomic descriptors capture texture patterns not trivially
explained by global ROI size, particularly in bone.

The OA classification results reinforce this interpretation. Across both
OAIZIB-CM and Po-OA, radiomics-based models consistently outperformed
models restricted to volume/thickness-linked features, with AUC gains of
roughly 0.1 or more across classifiers. Moreover, LASSO-selected
features from the full radiomics sets were only weakly correlated with
ROI size, whereas features selected from the size-linked pools remained
strongly size-dependent and yielded inferior discrimination. Together,
these findings indicate that our automated pipeline exposes meaningful
texture information beyond simple morphometric changes, even when using
routine clinical sequences rather than quantitative T\textsubscript{2} or T\textsubscript{1$\rho$}
maps. In contrast to studies that integrate only a few radiomic features
with clinical variables or deep features \cite{cui2023development, fu2025mri, chen2025integration, angelone2025innovative}, our work emphasises the
interplay between segmentation robustness, anatomical QC and feature
selection in determining the reliability and interpretability of OA
radiomics.

Several limitations should be acknowledged. First, although we evaluated
zero-shot behaviour on SKI-10 and Po-OA, all models were trained on
DESS-like OAIZIB-CM and SKM-TEA images; performance on other field
strengths, vendors or sequences remains to be established. Second, the
Po-OA cohort lacks voxelwise ground truth and uses report-derived OA
labels, which are noisy proxies rather than radiographic or arthroscopic
standards. Third, our radiomics analysis is cross-sectional and focuses
on binary OA classification; we did not investigate progression, pain
trajectories or response to interventions. Fourth, we used hand-crafted
radiomic features and classical machine-learning classifiers; more
advanced approaches such as graph neural networks or end-to-end
multi-task models might further improve performance but would also
require larger, harmonised datasets. Finally, although we excluded shape
features due to robustness concerns, we did not systematically assess
test--retest repeatability or the impact of different discretisation
schemes.

Moreover, our radiomics classification experiments were deliberately restricted to within-cohort training and validation, and we did not attempt to construct a cross-centre deployable OA classifier. Given the known sensitivity of hand-crafted radiomic features to acquisition protocols and scanners, robust multi-centre generalisation will likely require dedicated harmonisation or end-to-end deep learning approaches.

Future work will address these limitations by integrating longitudinal
data, harmonising multi-centre acquisitions, and exploring hybrid models
that combine LM-CartSeg ROIs with deep radiomic representations. The
proposed geometric L/M rule and QC framework are generic and could be
extended to other joints or imaging modalities, providing a reusable
backbone for anatomically grounded radiomics in musculoskeletal imaging.

\section{Conclusion}
We developed LM-CartSeg, a fully automated pipeline that couples robust
multi-structure segmentation with geometric L/M partitioning,
anatomical QC and radiomics analysis of knee MRI. Across public and
clinical cohorts, the method produced accurate cartilage and subchondral
bone masks, improved surface-based metrics via simple post-processing
rules, and delivered deterministic L/M compartments that remained stable
under domain shift.

Radiomics experiments showed that tibiofemoral cartilage--bone units
share strong texture signatures, that most radiomic features are not
simple surrogates of ROI size, and that OA vs.\ non-OA classification
benefits substantially from texture-based features beyond volume and
thickness. These results suggest that LM-CartSeg can serve as a
practical foundation for large-scale, reproducible radiomics studies in
knee osteoarthritis and may facilitate future work on early disease
detection and progression prediction in routine clinical MRI.

\section*{Data and Code Availability}
The public datasets used in this study are available from their original sources.
The OAIZIB-CM dataset can be accessed through the Osteoarthritis Initiative–ZIB collaboration under the terms specified by the data providers.
The SKI-10 dataset is publicly available as part of the MICCAI SKI-10 Grand Challenge.

The Po-OA dataset consists of retrospective clinical knee MRI scans collected under institutional ethical approval: HSEARS20180110001. Due to patient privacy, data protection regulations, and the conditions of informed consent, these data cannot be made publicly available. Access may be granted to qualified researchers upon reasonable request, subject to ethical approval, data use agreements, and approval by the corresponding institutional review board.

\section*{Ethics Statement}
This study was conducted in accordance with the Declaration of Helsinki. The publicly available datasets: OAIZIB-CM and SKI-10, were used under their respective data use terms and involved secondary analysis of de-identified data. The private Po-OA cohort was collected under institutional review board approval: HSEARS20180110001, all data were anonymized prior to analysis and no identifiable personal information was accessed.

\section*{Declaration of competing interest}

\bibliographystyle{elsarticle-num-names}
\bibliography{refs}
\end{document}